**Learning and inference in knowledge-based probabilistic model for medical diagnosis**

**Jingchi Jiang[a], Chao Zhao[a], Yi Guan[a,*], Qiubin Yu[b]**

[a] **School of Computer Science and Technology, Harbin Institute of Technology, Harbin 150001, China**

[b] **Medical Record Room, The 2nd Affiliated Hospital of Harbin Medical University, Harbin 150086, China**

\* **Correspondence address: Yi Guan, School of Computer Science and Technology, Harbin Institute of Technology, Comprehensive Building 803, Harbin Institute of Technology, Harbin 150001, China. Tel.: +86-186-8674-8550.**

*E-mail addresses:* **guanyi@hit.edu.cn (Y. Guan), jiangjingchi0118@163.com (J.C. Jiang), hitsa.zc@gmail.com (C. Zhao), yuqiubin6695@163.com (Q.B. Yu).**

# Learning and inference in knowledge-based probabilistic model for medical diagnosis

Jingchi Jiang[a], Chao Zhao[a], Yi Guan[a,*] Qiubin Yu[b]

[a] School of Computer Science and Technology, Harbin Institute of Technology, Harbin 150001, China

[b] Medical Record Room, The 2nd Affiliated Hospital of Harbin Medical University, Harbin 150086, China

*    Correspondence address: Yi Guan, School of Computer Science and Technology, Harbin Institute of Technology, Comprehensive Building 803, Harbin Institute of Technology, Harbin 150001, China. Tel.: +86-186-8674-8550.

*E-mail addresses:* guanyi@hit.edu.cn (Y. Guan), jiangjingchi0118@163.com (J.C. Jiang), hitsa.zc@gmail.com (C. Zhao), yuqiubin6695@163.com (Q.B. Yu).

**Abstract**

Based on a weighted knowledge graph to represent first-order knowledge and combining it with a probabilistic model, we propose a methodology for the creation of a medical knowledge network (MKN) in medical diagnosis. When a set of symptoms is activated for a specific patient, we can generate a ground medical knowledge network composed of symptom nodes and potential disease nodes. By Incorporating a Boltzmann machine into the potential function of a Markov network, we investigated the joint probability distribution of the MKN. In order to deal with numerical symptoms, a multivariate inference model is presented that uses conditional probability. In addition, the weights for the knowledge graph were efficiently learned from manually annotated Chinese Electronic Medical Records (CEMRs). In our experiments, we found numerically that the optimum choice of the quality of disease node and the expression of symptom variable can improve the effectiveness of medical diagnosis. Our experimental results comparing a Markov logic network and the logistic regression algorithm on an actual CEMR database indicate that our method holds promise and that MKN can facilitate studies of intelligent diagnosis.

## 1. Introduction

The World Health Organization (WHO) reports that 422 million adults have diabetes, and 1.5 million deaths are directly attributed to diabetes each year [1]. Additionally, the number of deaths caused by cardiovascular diseases (CVDs) and cancer is estimated to be 17.5 million and 8.2 million, respectively [2]. The WHO report on cancer shows that new cases of cancer will increase by 70 percent over the next two decades. In the face of this situation, researchers have begun to pay more attention to health care. According to existing studies, more than 30 percent of cancer deaths could be prevented by early diagnosis and appropriate treatment [3]. Because an accurate diagnosis contributes to a proper choice of treatment and subsequent cure, medical diagnosis plays a great role in the improvement of health care. Consequently, a means to provide an effective intelligent diagnostic method to assist clinicians by reducing costs and improving the accuracy of diagnosis has been a critical goal in efforts to enhance the patient medical service environment.

Classification is one of the most widely researched topics in medical diagnosis. The general model classifies a set of symptom data into one of several predefined categories of disease for cases of medical diagnosis. A decision tree [4-5] is a classic algorithm in the medical classification domain, one that uses the information entropy method; however, it is sensitive to inconsistencies in the data. The support vector machine [6-8] has a solid theoretical basis for the classification task; because of its efficient selection of features, it has higher predictive accuracy than decision trees. Bayesian networks [9-10], which are based on Bayesian theory [11-12], describe the dependence relationship between the symptom variables and the disease variables; these can be used in medical diagnosis. Other diagnostic models include neural networks (NN) [13-15], fuzzy logic (FL) [16-17], and genetic algorithms (GAs) [18-20]. Each of these is designed with a distinct methodology for addressing diagnosis problems.

Existing studies have mainly focused on exploring effective methods for improving the accuracy of disease classification. However, these methods often ignore the importance of the application of domain knowledge. Although Markov logic network [21] is a probabilistic inferential model based on the first-order logic rules, it only applies to binary features which is against the numerical characteristic of symptom. In this paper, we focus on combining medical knowledge with a novel probabilistic model to assist clinicians in making intelligent diagnoses. We conducted our investigation as follows:

(1) In order to obtain medical knowledge from Chinese Electronic Medical Records (CEMRs), we adopted techniques for the recognition of named entities and entity relationships. By mapping named entities and entity relationships into sentences, we built a medical knowledge base consisting of a set of rules in first-order logic.

(2) We mapped the first-order knowledge base into a knowledge graph. This graph is composed of first-order predications (nodes) and diagnostic relationships among predications (edges). Furthermore, the graph can also be an intuitive reflection of the inferential structure of the knowledge.

(3) We developed a novel probabilistic model for medical diagnosis that is based on Markov network theory. For adapting to the requirements of multivariate feature in medical diagnosis, we incorporated a Boltzmann machine into the potential function of a Markov network. It can simultaneously model both binary and numerical indexes of symptoms. The mathematical derivation of learning and inference is rigorously deduced.

(4) By a numerical comparison with other diagnostic models for CEMRs, we found that our probabilistic model is more effective for diagnosing several diseases according to the measures of precision for the first 10 results (P@10), recall for the first 10 results (R@10), and average discounted cumulative gain (DCG-AVG).

The rest of this paper is organized as follows. In Section 2, we introduce Chinese Electronic Medical Records and the knowledge graph. In Section 3, we review the fundamentals of Markov networks and Markov logic networks. In Section 4, the knowledge-based probabilistic model based on Markov networks is proposed; then, we demonstrate the mathematical derivation of learning and inference. In Section 5, we further evaluate the effectiveness and accuracy of our probabilistic model for medical diagnosis. Finally, we conclude this paper and discuss directions for future work in Section 6.

## 2. Knowledge extraction and knowledge representation

### 2.1. Chinese Electronic Medical Records

Electronic medical records (EMRs) [22] are a systematized collection of patient health information in a digital format. As the crucial carrier of recorded medical activity, EMRs contain significant medical knowledge [23-24]. Therefore, for this study we adopted Chinese Electronic Medical Records (CEMRs) in free-form text as the primary source of medical knowledge. These CEMRs, which have had protected health information (PHI) [25] removed, come from the Second Affiliated Hospital of Harbin Medical University, and we have obtained the usage rights for research. These CEMRs mainly include five kinds of free-form text: discharge summary, progress note, complaints of the patient, disease history of the patient, and the communication log. Considering the abundance of medical knowledge and the difficulty of Chinese text processing, we chose the discharge summary and the progress note as the source for knowledge extraction. The structures of the discharge summary and progress note are shown in Figs. 1 and 2, respectively.

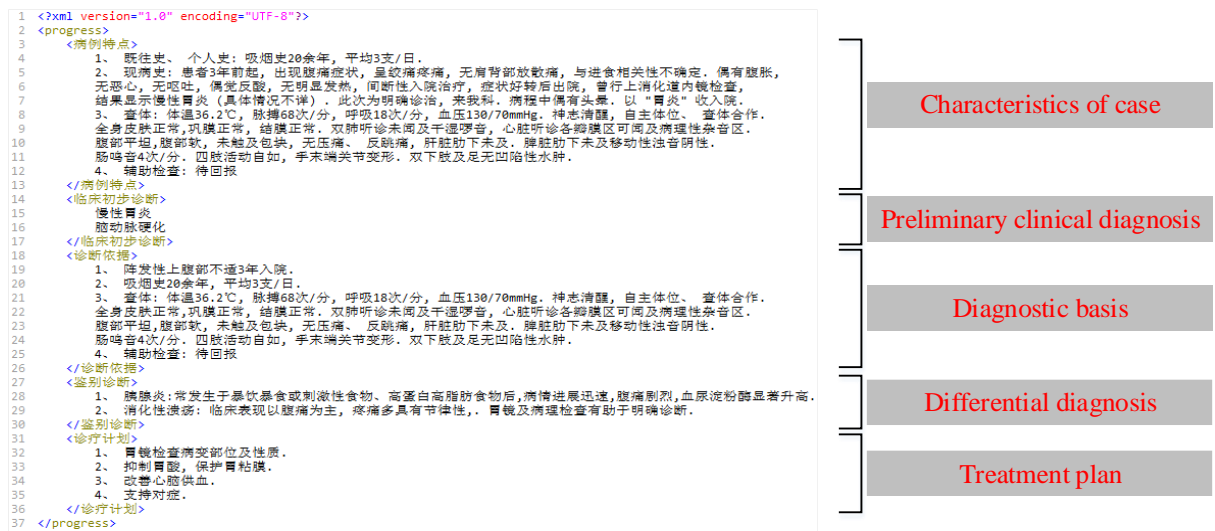

**Fig. 1.** Sample of discharge summary from the Second Affiliated Hospital of Harbin Medical University.

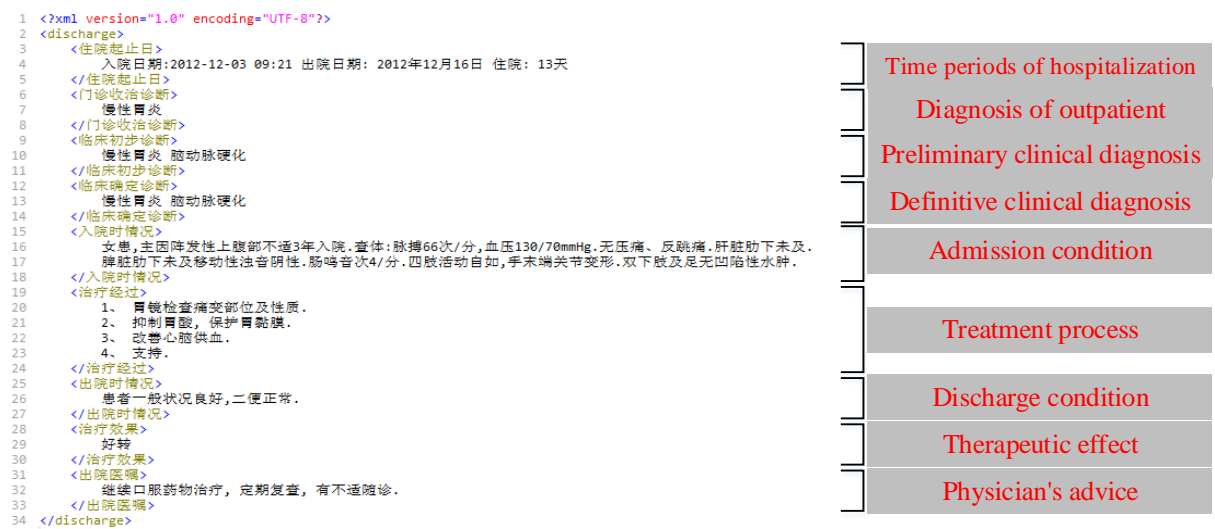

**Fig. 2.** Sample of progress note from the Second Affiliated Hospital of Harbin Medical University.

## 2.2. Corpus

The recognition of named entities [26] and entity relationships [27] is an important aspect in the extraction of medical knowledge from CEMRs. Referencing the medical concept annotation guideline and the assertion annotation guideline given by Informatics for Integrating Biology and the Bedside (i2b2) [28], we have drawn on the guidelines for CEMRs [29] and manually annotated the named entity and entity relationship of 992 CEMRs as the resource of medical knowledge. For this diagnostic task, this study only kept "symptom" entities, "disease" entities, and the "indication" relationship. The "indication" relationship holds when the related "symptom" indicates that the patient suffers from the related "disease." In addition, there are three modifiers for "symptom" entities, namely *present*, *absent*, and *possible*.

## 2.3. Knowledge graph

The medical knowledge obtained from the 992 CEMRs can be comprehended as a set of first-order logic rules among "symptom" entities, "disease" entities, and "indication" relationships. The reliability of medical knowledge corresponds to the probability of the "indication" relationship. By gathering all the annotated "indication" relationships, a medical knowledge base may be constructed. However, the medical knowledge base lacks the connectivity of real-world knowledge. In order to capture this medical knowledge more intuitively, we build a more comprehensive knowledge graph consisting of "disease" and "symptom" entities as nodes and the "indication" relationships as edges. As the reliability of medical knowledge increases, the corresponding edge's weight gradually grows. The topology of the knowledge graph is shown in Fig. 3.

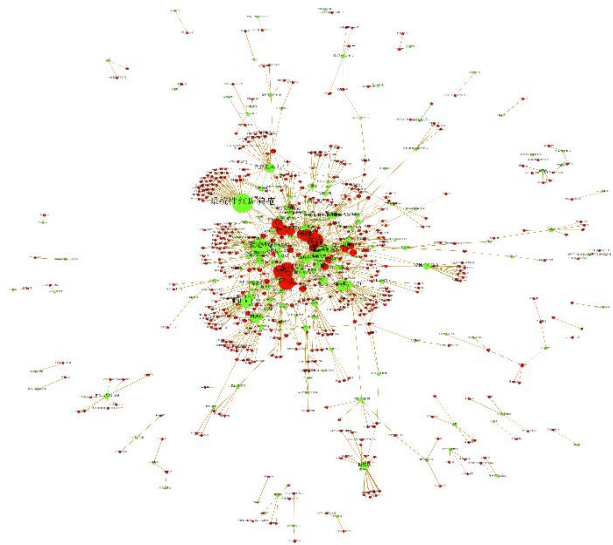

**Fig. 3.** Topology of the knowledge graph.

The nodes in the knowledge graph are divided into two different colors according to the type of entity, the red nodes and the green nodes representing "symptom" entities and "disease" entities, respectively. This graph contains 173 kinds of disease and 508 kinds of symptom. As a whole, 1069 pieces of knowledge are embodied in the knowledge graph.

## 3. Markov logic networks (MLNs)

As a uniform framework of statistical relational learning, a Markov logic network (MLN) combines first-order logic with a probability graph model for solving problems of complexity and uncertainty. From a probability-and-statistics point of view, MLN is based on the methodology of Markov networks (MNs) [30]. From a first-order-logic point of view, it can briefly present the uncertainty rules and can tolerate incomplete and contradictory problems in the knowledge areas.

### 3.1. Markov networks and first-order logic

A Markov network, which is a model for the joint distribution of a set of variables $X = (X_1, X_2, ..., X_n) \in \chi$, provides the theoretical basis for a Markov logic network. The Markov network is composed of an undirected graph $G$ and a set of potential functions $\phi_k$. The joint distribution of the Markov network is given as

$$P(X = x) = \frac{1}{Z} \prod_k \phi_k(x_{\{k\}}) \tag{1}$$

where $x_{\{k\}}$ is the state of the $k$th clique. $Z$, known as the normalization function, is given by $Z = \sum_{x \in \chi} \prod_k \phi_k(x_{\{k\}})$. The most widely used method for approximate inference in MN is Markov chain Monte Carlo (MCMC), and Gibbs sampling in particular. Another popular inference method in MN is the sum–product algorithm.

A medical knowledge base is a set of rules in first-order logic. Rules are composed using four types of symbols: constants, variables, functions, and predicates. A term is any expression representing an object. An atom is a predicate symbol applied to a tuple of terms. A ground atom is an atomic rule, all of whose arguments are ground terms. A possible world assigns a truth value to each possible ground atom.

### 3.2. Markov logic networks in medical diagnosis

A Markov logic network can be considered as a template for generating Markov networks. Given different sets of constants, it will generate different Markov networks. According to the definition of a Markov network, the joint distribution of a Markov logic network is given by

$$P(X = x) = \frac{1}{Z} \exp(\sum_i \omega_i n_i(x)) = \frac{1}{Z} \prod_i \phi_i(x_{\{i\}})^{n_i(x)} \tag{2}$$

where $n_i(x)$ is the number of true groundings of the $i$th rule $R_i$ in constants $x$; $\omega_i$ is the weight for $R_i$; $x_{\{i\}}$ is the state of the atoms appearing in $R_i$; and $\phi_i(x_{\{i\}}) = e^{\omega_i}$. Because MLN only focuses on binary features, the constants $x$ are discrete values and $x \in \{0,1\}$.

To apply MLN in medical diagnosis, the atom is considered as the medical entity. When the medical entity presents an explicit condition for a patient, the corresponding atom of this entity in MLN is assigned the value 1; otherwise, 0. By this mapping method, we are able to convert medical knowledge into the binary rules of the MLN. As medical knowledge accumulates, an MLN will be built, and the maximum probability model of MLN can be used for medical diagnosis. Given a series of symptoms, the risk probability for a specific disease is calculated by

$$\arg\max_{y} P(y \mid x) = \arg\max_{y} \sum_{i} \omega_i n_i(x, y) \tag{3}$$

Because of the higher complexity of calculating $n_i(x, y)$, the problem of maximum probability can be transformed into a satisfiability problem, for which a set of variables is searched to maximize the number of rules satisfied.

## 4. Methodology

Although MLN can be used for medical diagnosis, it is only suitable for binary rules. The reason is that the values of $n_i(x, y)$ are uncountable when $x$ is a continuous variable. Thus, MLN is inefficient for multivariate rules. In the health care field, the indexes of symptoms are often expressed in numeric form. Thus, the existing MLN methodology has some obvious shortcomings for numeric-based diagnosis. This section addresses that problem. By changing the form of expression of the potential function, we can incorporate the continuous variable $x$ into the joint distribution of MN, enabling the conditional probability model for inference to be deduced via Boltzmann machine, and the learning model for calculating the weight for each rule is proposed.

### 4.1. Medical knowledge network (MKN)

Based on the previously mentioned knowledge graph, we propose a model for handling numeric-based diagnosis, combining the knowledge graph with the theoretical basis of Markov networks. The novel theoretical framework is named the medical knowledge network (MKN).

**Definition 1.** *A medical knowledge network L is a set of pairs* $(R_i, \omega_i)$ *where* $R_i$ *is the medical knowledge in first-order logic and* $\omega_i$ *is the reliability of* $R_i$. *Together with a finite set of constants* $C = \{c_1, c_2, ..., c_n\}$, *it defines a ground medical knowledge network* $M_{L,C}$ *as follows:*

*1.* $M_{L,C}$ *contains one multivariate node for each possible grounding of each medical entity appearing in L. The value of the node is the quantified indicator of the symptom entity or disease entity.*

*2.* $M_{L,C}$ *contains one weight for each piece of medical knowledge. This weight is the* $\omega_i$ *associated with* $R_i$ *in L.*

In a Markov network, a potential function is a nonnegative real-valued function of the state of the corresponding clique. Therefore, the potential function of MKN can also be regarded as the state of a clique, which is composed of one or more "indication" relationships. Incorporating the quantified indicator of each entity into the potential function is an important step for numeric-based diagnosis. From statistical physics, we can express the potential function as an energy function [31], rewriting $\phi(D)$ as

$$\phi(D) = \exp(-\varepsilon(D)) \tag{4}$$

where $\varepsilon(D)$ is often called an energy function. The set $D$ is the state of the "symptom" entity and "disease" entity. Then, the expression $\varepsilon(D)$ is interpreted in terms of an unrestricted Boltzmann machine [32], which is the one of the earliest types of Markov network. The energy function associated with the "indication" relationship is defined by a particularly simple parametric form,

$$\varepsilon(D) = \varepsilon_{ij}(x_i, x_j) = -\omega_{ij} x_i x_j \tag{5}$$

where $x_i$ and $x_j$ represent the value of the "symptom" entity and the "disease" entity, respectively, and $\omega_{ij}$ is the contribution of the energy function. According to Eq. (1), the joint distribution is defined as follows:

$$
\begin{aligned}
P(X = x) &= \frac{1}{Z} \prod_{\substack{i \\ r_i \in R}} \phi_i(D) = \frac{1}{Z} \prod_{\substack{i \\ r_i \in R}} \exp(-\varepsilon(D)) \\
&= \frac{1}{Z} \exp\left( -\sum_{\substack{i \\ r_i \in R}} (-\omega_i x_{r_i}^s x_{r_i}^d) \right) \\
&= \frac{1}{Z} \exp\left( \sum_{\substack{i \\ r_i \in R}} \omega_i x_{r_i}^s x_{r_i}^d \right)
\end{aligned}
\tag{6}
$$

In general, the energy function of an unrestricted Boltzmann machine contains a set of parameters $u_i$ that encode individual node potentials. These activated individual variables will stress the effect of the "symptom" entity in the energy function. The rewritten probability formula is given as

$$P(X = x) = \frac{1}{Z} \exp\left( \sum_{\substack{i \\ r_i \in R}} (\omega_i x_{r_i}^s x_{r_i}^d + u_{x_{r_i}^s} x_{r_i}^s) \right) \tag{7}$$

As can be seen, when a "symptom" entity is activated, the factor of the corresponding individual node potential will be considered a major component of the model; this is exactly consistent with a clinical diagnosis that is based on symptoms. In this paper, we adopt the Gaussian potential function (GPF) as the individual node potential, which is expressed as

$$u_{x_{r_i}} = \sum_{j=1}^{n} \left( m_{x_j^d} e^{-(\frac{d_{x_{r_i}^s x_j^d}}{\sigma})^2} \right) \tag{8}$$

where $d_{x_{r_i}^s x_j^d}$ represents the distance between node $x_{r_i}^s$ and its neighboring node $x_j^d$ in the knowledge graph. The influence factor $\sigma$ is used for controlling the influence range of each node and $m_{x_j^d}$ is the quality of node $x_j^d$. The final probability model is defined as the following distribution:

$$P(X = x) = \frac{1}{Z} \exp\left( \sum_{\substack{i \\ r_i \in R}} \left( \omega_i x_{r_i}^s x_{r_i}^d + \left( \sum_{j=1}^{n} m_{x_j^d} e^{-(\frac{d_{x_{r_i}^s x_j^d}}{\sigma})^2} \right) \cdot x_{r_i}^s \right) \right) \tag{9}$$

Using Definition 1 and the deduced joint distribution, Algorithm 1 provides the procedure for building a medical knowledge network.

**Algorithm 1.** Construction of medical knowledge network

---

**Input:** $List_{EMR}$: a list of the electronic medical records for training.
**Output:** $Network$: a medical knowledge network.
***Begin***
1:  Initialize the lists of nodes $Set_{node}$ and list of edges $Set_{edge}$ in MKN.
2:  Extract the entities and relationships from the $List_{EMR}$ → $Rules = \{rule_1,$ $rule_2, \ldots, rule_n\}$.
3:  Initialize the weights for $Rules$ by a fixed value $\omega$.
4:  ***for*** $rule_i \in Rules$ ***do***
5:      Initialize the symptom node $Node_{symptom}$ and disease node $Node_{disease}$.
6:      Parse the symptom predicate and the disease predicate from $rule_i$ → $Node_{symptom}$ and $Node_{disease}$.
7:      $Set_{node} \leftarrow Node_{symptom}$.
8:      $Set_{node} \leftarrow Node_{disease}$.
9:      Define the relationship $Edge_i$ between $Node_{symptom}$ and $Node_{disease}$.
10:     Add $Edge_i$ to $Set_{edge}$, and assign $\omega$ as the weight of $Edge_i$.
11: ***end for***
12: ***Function*** $PageRank(Set_{node}, Set_{edge})$ ***end***
13: After calculating the PageRank of all nodes, the MKN is built: $Set_{node}, Set_{edge}$ → $Network$.
14: ***return*** $Network$.

---

Through traversing the rules and parsing the predicates, a medical knowledge network can be implemented. The PageRank function is used as the quality of each node. To characterize the reliability of medical knowledge, we set up a fixed value $\omega$ for the initial network.

## 4.2. Inference

Medical inference can answer the two common generic clinical questions: "What is the probability that rule $R_1$ holds given rule $R_2$?" and "What is the probability of disease $D_1$ given the symptom vector $S_1$?" In response to the first problem, we can answer by computing the conditional probability as

$$
\begin{aligned}
P(R_1 \mid R_2, L, C) &= P(R_1 \mid R_2, M_{L,C}) \\
&= \frac{P(R_1 \wedge R_2 \mid M_{L,C})}{P(R_2 \mid M_{L,C})} \\
&= \frac{\sum_{x \in \chi_{R_1} \cap \chi_{R_2}} P(X = x \mid M_{L,C})}{\sum_{x \in \chi_{R_2}} P(X = x \mid M_{L,C})}
\end{aligned}
\tag{10}
$$

The set $\chi_{R_i}$ is the set of rules where $R_i$ holds, and $P(x \mid M_{L,C})$ is given by Eq. (9). Through the free combinations of pairs of disconnected atoms in MKN, some new rules will be derived by inference. When the probability of a new rule exceeds a certain threshold, it can be concluded that the new rule is reliable under the current base. Rule inference not only helps enrich the knowledge base but is also a self-learning mechanism for the MKN.

The second inference question is what is usually meant by "disease diagnosis." On the condition that the patient has a given symptom vector, we can predict the risk probability for a specific disease. This can be classified as a typical problem of conditional probability. The risk probability of disease $y$ can be calculated by

$$
\begin{aligned}
&P(Y = y \mid B_l = b_l) \\
&= \frac{\exp\left( \sum_{r_i \in R_l}^{i} \left( \omega_i x_{r_i} y_{r_i} + \left( \sum_{j=1}^{n} m_{y_j} e^{-\left(\frac{d_{x_0 y_j}}{\sigma}\right)} \right) \cdot x_{r_i} \right) \right)}{\exp\left( \sum_{r_i \in R_l}^{i} \left( \omega_i x_{r_i} y_{r_i}^{0} + \left( \sum_{j=1}^{n} m_{y_j} e^{-\left(\frac{d_{x_0 y_j}}{\sigma}\right)} \right) \cdot x_{r_i} \right) \right) + \exp\left( \sum_{r_i \in R_l}^{i} \left( \omega_i x_{r_i} y_{r_i}^{1} + \left( \sum_{j=1}^{n} m_{y_j} e^{-\left(\frac{d_{x_0 y_j}}{\sigma}\right)} \right) \cdot x_{r_i} \right) \right)}
\end{aligned}
\tag{11}
$$

where $R_l$ is the set of ground rules in which disease $y$ appears, and $b_l$ is the Markov blanket of $y$. The Markov blanket of a node is the minimal set of nodes that renders it independent of the remaining network; this is simply the set of that node's neighbors in the knowledge graph. Corresponding to the $i$th ground rule, $y_{r_i}$ is the value (0 or 1) of disease $y$. In contrast with the MLN diagnostic model, MKN avoids the complexity problem of $n_i(x, y)$ and incorporates the quantitative value of symptom $x_{r_i}$ into the diagnostic model. The detailed diagnostic algorithm is shown in Algorithm 2.

**Algorithm 2.**   Disease diagnostic algorithm based on MKN

---

**Inputs**: *Rules*: a set of rules with the learned weights   $\omega$.
           *PR*: a set of PageRank values for the nodes in MKN.
           *Evidences*: a set of ground atoms with known values for a specific patient.
           *Query*: a set of ground atoms with unknown disease values.
**Output**: *Result*: a diagnosis result for the specific patient.
***Begin***
1:  **for**   $disease_i \in Query$   **do**
2:         Initialize the probability $Pro_{activated}$ with the activated $disease_i$.
3:         Initialize the probability $Pro_{inactivated}$ with the inactivated $disease_i$.
4:         //Activating the disease atoms in *Network*.
5:         **for**   $rule_j \in Rules$   **do**
6:                $Pro_{inactivated} \mathrel{+}= \omega_j \cdot symptom_j \cdot disease_j + PR_{disease_j} \cdot symptom_j/E$.
7:                **if**   $disease_i \in rule_j$
8:                       Activate the atom of $disease_j$.
9:                **end if**
10:                $Pro_{activated} \mathrel{+}= \omega_j \cdot symptom_j \cdot disease_j + PR_{disease_j} \cdot symptom_j/E$.
11:         **end for**
12:         $Result_i = exp(Pro_{activated})/(exp(Pro_{activated}) + exp(Pro_{inactivated}))$.
13: **end for**
14: ***Function***   $Sort(Result)$   ***end***
15: **return**   $Result$.

---

Following the Eq. (11), we propose a disease diagnostic algorithm based on MKN. To provide a reliable diagnosis, we need calculate the risk of each disease. According to the evidences, Algorithm 2 can generate a list of potential disease which is sorted by diagnostic possibility.

### 4.3. Learning

A learning model is proposed for the calculation of the weight for each piece of medical knowledge from the Chinese Electronic Medical Records. In this study, we adopted the gradient descent method. Assuming independence among diseases, the learning model first calculates the joint probability distribution of a disease vector:

$$P_\omega^*(Y = y) = \prod_{l=1}^{m} P_\omega\left(Y_l = y_l \mid M_{L,C}\right) \tag{12}$$

where $m$ is the dimension of disease vector $y$. By Eq. (12), the derivative of the log-likelihood function with respect to the weight for the $i$th rule is

$$\begin{aligned}
\frac{\partial}{\partial \omega_i} \log P_\omega^*(Y = y) &= \frac{\partial}{\partial \omega_i} \log \prod_{l=1}^{m} P_\omega\left(Y_l = y_l \mid M_{L,C}\right) \\
&= \sum_{l=1}^{m} \frac{\partial}{\partial \omega_i} \log P_\omega\left(Y_l = y_l \mid M_{L,C}\right)
\end{aligned} \tag{13}$$

The calculation of $\dfrac{\partial}{\partial \omega_i} \log P_\omega\left(Y_l = y_l \mid M_{L,C}\right)$ will be a stubborn problem. Therefore, we try to construct the derivative of the log-likelihood. From Eq. (9), we know that the normalization function $Z$ can be expressed as

$$Z = \sum_{y \in \eta} \exp\left(\sum_{\substack{i \\ r_i \in R}} \left(\omega_i x_{r_i} y_{r_i} + \left(\sum_{j=1}^{n} m_{y_j} e^{-\left(\frac{d_{x_{r_i}y_j}}{\sigma}\right)}\right) \cdot x_{r_i}\right)\right) \tag{14}$$

Then, we have the pseudo-log-likelihood of Eq. (9) and its gradient:

$$\log P(Y = y \mid M_{L,C}) = \sum_{\substack{i \\ r_i \in R}} \left(\omega_i x_{r_i} y_{r_i} + \sum_{j=1}^{n} \left(m_{y_j} e^{-\left(\frac{d_{x_{r_i}y_j}}{\sigma}\right)}\right) \cdot x_{r_i}\right) - \log Z \tag{15}$$

$$\begin{aligned}
\frac{\partial}{\partial \omega_i} \log P(Y = y \mid M_{L,C}) &= x_{r_i} y_{r_i} - \frac{1}{Z}\left(\sum_{y \in \eta} \exp\left(\sum_{\substack{i \\ r_i \in R}} \left(\omega_i x_{r_i} y_{r_i} + \left(\sum_{j=1}^{n} m_{y_j} e^{-\left(\frac{d_{x_{r_i}y_j}}{\sigma}\right)}\right) \cdot x_{r_i}\right)\right)\right) \cdot x_{ri} y_{r_i} \\
&= x_{r_i} y_{r_i} - \sum_{y \in \eta} P(Y = y \mid M_{L,C}) \cdot x_{r_i} y_{r_i}
\end{aligned} \tag{16}$$

where $\eta$ is the set of all possible values of $y$, and $P(Y = y \mid M_{L,C})$ can be given by Eq. (9).

By bringing Eq. (16) into Eq. (13), the derivative of the log-likelihood with respect to the weight for the $i$th rule can be naturally calculated. We get the final expression

$$\frac{\partial}{\partial \omega_i} \log P_\omega^*(Y = y) = \sum_{l=1}^{m} \left( x_{r_i} y_{r_i} - \sum_{y \in \eta} P(Y = y \mid M_{L,C}) \cdot x_{r_i} y_{r_i} \right) \tag{17}$$

After finite iterations, $\omega_i$ is calculated with the learning rate $\lambda$.

$$\omega_{i,t} = \omega_{i,t-1} + \lambda \frac{\partial}{\partial \omega_i} \log P_\omega^*(Y = y) \mid \omega_{t-1} \tag{18}$$

The detailed procedure for the weight learning model is presented in Algorithm 3.

**Algorithm 3.** Learning algorithm for MKN

---

**Inputs**: *Network*: an MKN with vector $\omega$ of fixed weights.
        *Evidences*: a set of ground atoms with known values.
**Output**: *Weights*: a learned weight vector.
***Begin***
1:   Initialize the weight vector *Weights*.
2:  ***for*** *weight$_i$* ∈ *Weights* ***do***    *//weight$_i$* represents the weight for the *i*th rule
3:     ***while*** *t From* 1 *To* 100 ***do***
4:       ***for*** *evidence$_j$* ∈ *Evidences* ***do***   *//*evidence set for the *j*th patient
5:          Extract the blanket of the *i*th rule → *blanket$_i$*.
6:          *//*Mapping *evidence$_j$* to *blanket$_i$*
7:          *slope* += $s_{ij} \cdot d_{ij} - \sum_{x' \in X}[P_\omega(X_l = x' | blanket_i) \cdot s_{ij} \cdot d_{ij}]$.
8:          *//s$_{ij}$* represents the symptom value of *evidence$_j$* for the *i*th rule
9:          *//d$_{ij}$* represents the disease value of *evidence$_j$* for the *i*th rule
10:      ***end for***
11:      $\omega_{i,t} = \omega_{i,t-1} + \eta \cdot slope$.
12:    ***end while***
13:    *weight$_i$* ← $\omega_{i,t}$.
14: ***end for***
15: ***return*** *Weights*.

---

In summary, we adopt the log-likelihood function and the gradient descent method to learn the weight vector. Fortunately, the gradient of pseudo-log-likelihood can be calculated by the joint probability distribution in finite time. Mapping the evidence to its Markov blanket is also to improve the time-effectiveness of learning algorithm.

## 5. Experiments and discussion

In order to verify the effectiveness of the medical knowledge network, we conducted experiments using actual CEMRs. Based on the knowledge graph concept described in Section 2.2, we built an MKN for medical diagnosis. We chose the manually annotated 992 CEMRs with the help of medical professionals and only kept the discharge summary and the progress note as the source of knowledge. In the annotating process, we classified the entities into five categories: disease, type of disease, symptom, test, and treatment; only the disease entity and the symptom entity were extracted to complete the diagnostic task. Additionally, owing to the lack of numerical indexes for symptoms in our CEMRs, we adopted modifiers for the symptom, namely *present*, *possible*, and *absent*, to represent the symptom variable $x$, corresponding to 2, 1, and 0, respectively. Although the modifier of the symptom is not a continuous variable, a multivariate version of MKN also has theoretical significance.

After the MKN was constructed, we randomly selected 300 untagged CEMRs as the test corpus, and conditional random fields (CRFs) were used to automatically recognize the disease entities and symptom entities. Based on the symptom entities on each CEMR, we inferred the diagnosis result and ascertained whether there was consistency between the diagnosed disease and the actual disease.

The description and analysis of the experiments are mainly concerned with three aspects: the parameter analysis, the weight learning, and the relative effectiveness of MKN and the other methods compared.

### 5.1. Parameter analysis

In this section, we focus on the optimum choices for the parameter values. In Eq. (11), $d_{x_{r_i} y_j}$ is the distance between $x_{r_i}$ and its neighboring node $y_j$ in the knowledge graph. Therefore, we define $d_{x_{r_i} y_j} = 1$. The influence factor $\sigma$ represents the control range of each node. If a symptom atom and a disease atom appear in a common rule, they have an interaction with each other, and the two atoms in each rule can be represented as two adjacent nodes in the knowledge graph. Since we naturally assume that the symptom node only affects the nearest connected disease node, we set $\sigma = 1$.

The terms $m_{y_j}$, which is the quality of the disease node, and $x_{r_i}$, which is the symptom variable in the $i$th rule, are both uncertainty parameters. The selection of expressions for $m_{y_j}$ and $x_{r_i}$ will affect the accuracy of MKN in diagnosing disease. To begin, we experimented with three classical measures for $m_{y_j}$: PageRank, degree, and betweenness centrality, and we used discounted cumulative gain (DCG) [33] as the indicator to measure the accuracy of the diagnosis result. The DCG score can be calculated by

$$DCG_P = rel_1 + \sum_{i=2}^{P} \frac{rel_i}{\log_2^i} \tag{19}$$

where $rel_i$ represents the relevance of the $i$th disease in the diagnosis result; a correct diagnosis is 1, whereas a misdiagnosis is 0. The variable $P$ is the number of diagnosis results, which in this study was 10.

As structural differences between the discharge summary and the progress note can lead to different numbers of symptom for the same patient, two experiments were used to distinguish between them. Fig. 4 shows the DCG scores (y-axes) plotted against the serial numbers of 40 discharge summaries and 260 progress notes (x-axes) for the three measures of quality for the disease node.

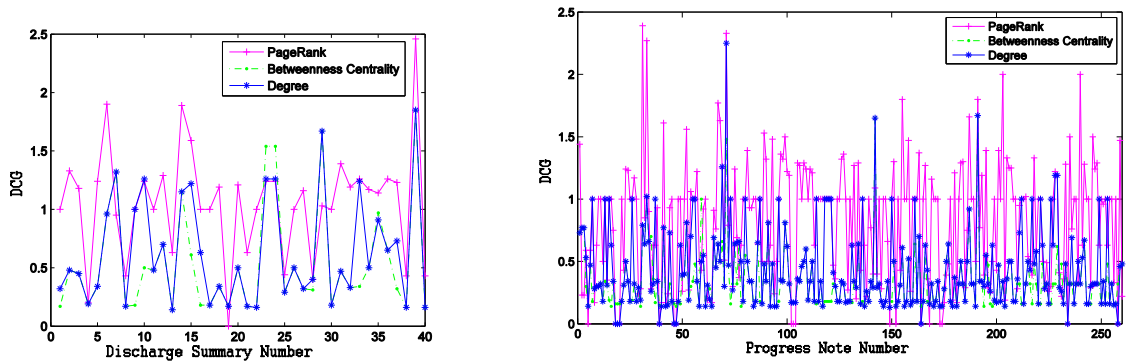

**Fig. 4.** DCG (discounted cumulative gain) for discharge summaries and progress notes using different measures of quality for the disease node.

Although the results show that the curves of the DCG scores are irregular, the DCG score is 1 in most cases. From the DCG descriptions, the reason is that most of the CEMRs have only one actual disease, and our model ranks this disease at the top of the diagnosis result. We also observe that the effectiveness of the PageRank-based MKN is better than the other methods for both the discharge summary and the progress note.

In order to describe the diagnosis result more directly, we adopt a second measure, R@10, which is the recall for the first 10 results. If $m$ actual diseases appear in a CEMR and the MKN returns $n$ of them, then the R@10 is given by

$$R@10 = \frac{n}{m} \qquad 0 \le n \le m, n \le 10 \qquad (20)$$

Fig. 5 shows the distributions of R@10 for discharge summaries and progress notes, with the blue, green, and red bars showing the results using PageRank, betweenness centrality, and degree, respectively. Under PageRank, the recall for nearly half the records is 1.0. By contrast, the results for betweenness centrality and degree are unsatisfactory because they have higher proportions with a recall of 0.0 and lower proportions with 1.0. Considering both factors DCG and R@10, we conclude that PageRank is more suitable to use as the quality of disease node

$m_{y_j}$ .

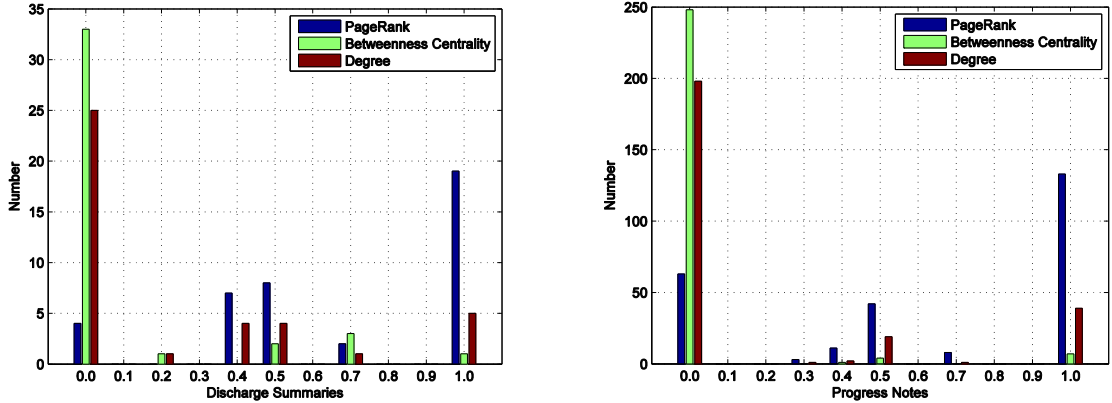

**Fig. 5.** Distribution of R@10 (recall for first 10 results) for discharge summaries and progress notes using different measures of quality for the disease node.

The second uncertainty parameter is $x_{r_i}$, which is the quantitative value of the symptom. Although the discrete modifier of the symptom could be employed as the representation of $x_{r_i}$ in this paper, it would not be the best choice for processing continuous values in the future. If the continuous value of a symptom is used directly as $x_{r_i}$, the problem of normalization across different symptoms will be an important factor that could cause undesirable results. Therefore, we seek a representation of $x_{r_i}$ that not only satisfies the requirements for discrete values but also might be suitable for continuous values. The sigmoid function is a typical normalization method. However, the domain of the sigmoid function does not match the value range for symptoms. For the diagnostic task, we designed an improved sigmoid function to express $x_{r_i}$, the quantitative value of symptom. The improved sigmoid function is defined as follows:

$$S(x) = \frac{2}{1 + e^{-(x - x_{normal})^2}} - 1 \qquad (21)$$

where $x$ is the value of the symptom, and $x_{normal}$ is the normal value, corresponding to *absent* (and represented by 0) in this paper. By the characteristic of a sigmoid function, we can map the symptom variable to a normalization interval, which is $0 \le S(x) < 1$.

To further answer what kinds of representation of $x_{r_i}$ can improve the accuracy of the diagnosis results, we continued with experiments comparing the sigmoid function, our improved sigmoid, and discrete modifiers of the symptom. The experimental results are shown in Figs. 6 and 7.

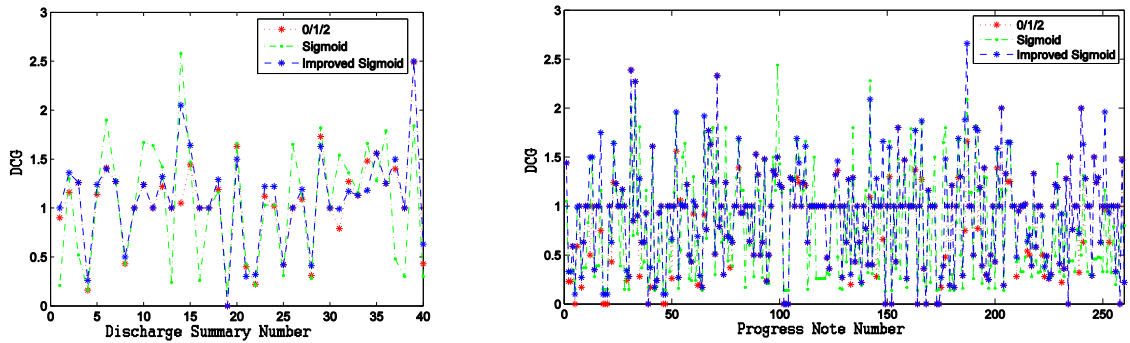

**Fig. 6.** DCG (discounted cumulative gain) for discharge summaries and progress notes using different types of symptom variable.

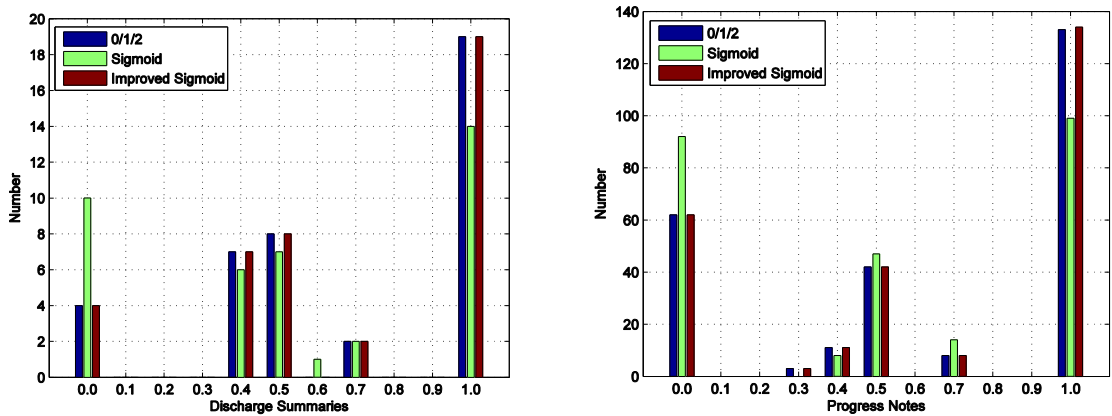

**Fig. 7.** Distribution of R@10 (recall for first 10 results) for discharge summaries and progress notes using different types of symptom variable.

Following the same evaluation criteria, the results shown in Fig. 7 indicate that the performance of the modifier-based variable is consistent with that of the improved sigmoid function, whereas in Fig. 6 the performance of the improved sigmoid function is shown to be a little better than that of the other methods. Hence, we conclude that the improved sigmoid function can not only handle the continuous symptom variable, but also performs well in the discrete field. In summary, considering MKN's ability to migrate between continuous variables and discrete variables, the improved sigmoid function should be employed as the expression of symptom variable $x_{r_i}$.

## 5.2. Weight learning

In this section, we make a credibility assumption: If a ground atom is in the knowledge base, it is assumed to be true; otherwise, it is false. In other words, the inference of the MKN depends completely on the existing medical knowledge. To test the effectiveness of the learning method, we compared four types of weighting, including constant weighting, MLN-based weighting, nonnegative MLN-based weighting, and MKN-based weighting. We divided MLN-based weight learning into typical weighting and nonnegative weighting. Because negative weights would be generated by MLN, violating the credibility assumption of medical knowledge, we rewrote the learning program "Tuffy," [34] which is an open-source MLN inference engine, to ensure that the learned weights would be nonnegative. When a negative weight is learned, our solution is to replace the negative weight with the current minimum positive weight at each iteration. In addition, we experimented using different constants as weights to check whether it might influence the diagnosis results. Tables 1 and 2 summarize the results for the discharge summaries and the progress notes, respectively, showing P@10, precision for the first 20 results (P@20), R@10, and average DCG.

**Table 1**

Analysis of effectiveness of weight learning for discharge summaries.

| Weight Type | P@10 | P@20 | R@10 | DCG-AVG |
|---|---|---|---|---|
| Constant Weight of 0.5 | 0.875 | 0.9 | 0.62 | 1.06 |
| Constant Weight of 1 | 0.875 | 0.9 | 0.62 | 1.06 |
| MLN Weight | 0.8056 | 0.8611 | 0.5233 | 0.822 |
| Positive MLN Weight | 0.8485 | 0.9091 | 0.51 | 0.8402 |
| MKN Weight | 0.9 | 0.95 | 0.67 | 1.0983 |

P@10 = precision for first 10 results; P@20 = precision for first 20 results; R@10 = recall for first 10 results; DCG-AVG = average discounted cumulative gain.
MLN = Markov logic network; MKN = medical knowledge network.

**Table 2**

Analysis of effectiveness of weight learning for progress notes.

| Weight Type | P@10 | P@20 | R@10 | DCG-AVG |
|---|---|---|---|---|
| Constant Weight of 0.5 | 0.7538 | 0.8115 | 0.5949 | 0.7909 |
| Constant Weight of 1 | 0.7538 | 0.8115 | 0.5949 | 0.7909 |
| MLN Weight | 0.6473 | 0.7593 | 0.5006 | 0.6743 |
| Positive MLN Weight | 0.7409 | 0.8455 | 0.5442 | 0.7348 |
| MKN Weight | 0.7615 | 0.8692 | 0.6337 | 0.8269 |

P@10 = precision for first 10 results; P@20 = precision for first 20 results; R@10 = recall for first 10 results; DCG-AVG = average discounted cumulative gain.
MLN = Markov logic network; MKN = medical knowledge network.

We can see that the constant weights of 0.5 and 1 give exactly the same results, demonstrating that the diagnosis results are not at all influenced by the weights' being equally adjusted. Furthermore, the positive MLN-based weighting is better than the typical MLN-based weighting by all evaluation criteria, whether from discharge summaries or progress notes. This indicates that the results of diagnosis are significantly improved by positivizing the negatively weighted knowledge, further demonstrating the significance of negatively weighted knowledge for the reliability of our medical knowledge. Finally, we experimentally conclude that the weight learning methods in order of effectiveness are MKN-based, constant, positive MLN-based, and MLN-based.

**5.3. Comparison with other algorithms**

After determining the uncertainty parameters and the type of weights, we compared three diagnostic systems: MLN, MKN, and the logistic regression algorithm (LR). Fig. 8 shows the DCG curves and the distribution of R@10 using all CEMRs. MKN is clearly more accurate than the other methods, demonstrating the promise of this approach. According to the DCG scores, LR performs well in some CEMRs but very poorly in others; its recall values are uniformly poor. Although LR is used in diagnosing single diseases in some studies, it hardly applies to diagnosis of multiple diseases, especially when a patient's symptoms are sparse. Compared with LR, MLN performs better in the DCG, even surpassing MKN for some records. However, there is a greater difference between MLN and MKN in R@10. We believe that the theoretical mechanism of MLN being based on rough binary logic is the main cause of the poor effect; the binary atoms cannot precisely capture the degree of seriousness of the symptoms. As a result of this defect, the actual top diseases cannot be ranked in the top 10 to the extent possible, although these diseases are indeed detected, which is reflected in the DCG. On the other hand, this testifies to the advantage of MKN's multivariate atoms for medical diagnosis.

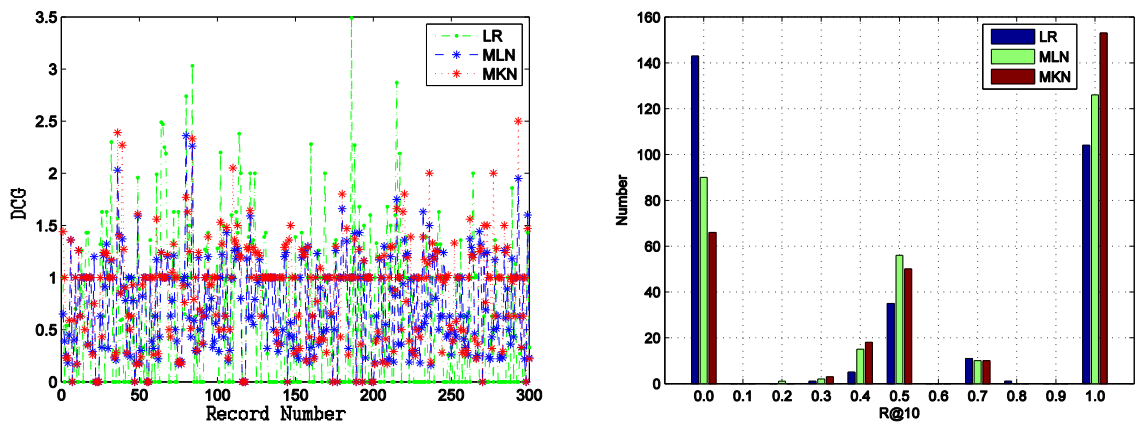

**Fig. 8.** DCG and R@10 for all CEMRs (Chinese Electronic Medical Records): MLN (Markov logic network), MKN (medical knowledge network), and LR (logistic regression algorithm).

Further, we calculated the average DCG and R@10 values for each of the three algorithms, which are shown in Table 3.

**Table 3**

Comparison of three diagnostic algorithms.

| Algorithm \ Index | DCG-AVG | R@10-AVG |
|---|---|---|
| LR | 0.6501 | 0.4383 |
| MLN | 0.7352 | 0.5548 |
| MKN | 0.8631 | 0.6385 |

DCG-AVG = average discounted cumulative gain; R@10-AVG = average values of recall for first 10 results.
LR = logistic regression algorithm; MLN = Markov logic network; MKN = medical knowledge network.

Overall, the best-performing diagnostic method is MKN, but about 20 percent of CEMRs are still misdiagnosed completely. The most likely reason is that the medical knowledge base created from 992 annotated CEMRs is minuscule. With the accumulation of medical knowledge, we believe that the usefulness of MKN as an intelligent system will continue to develop.

## 6. Conclusion and future work

In this paper, we have presented a knowledge-based probabilistic model for medical diagnosis. By extracting medical knowledge from Chinese Electronic Medical Records, a knowledge graph was constructed, which is composed of "disease" nodes and "symptom" nodes. Building on the theory of Markov networks and Markov logic networks, we developed a novel probabilistic model, called the medical knowledge network. In order to address the problem of numeric-based diagnosis, the model applies the energy function of Boltzmann machines as the potential function. Then, the mathematical derivation process of learning and inference were rigorously deduced. In contrast to a Markov logic network, the medical knowledge network adopts ternary rules or even continuously numeric rules, not being limited to binary rules. In experiments, PageRank and our improved sigmoid function were applied as the quality of disease node and the expression of the symptom variable, respectively. Empirical tests with actual records illustrate that MKN can improve diagnostic accuracy. Through comparisons with other algorithms, the effectiveness and promise of MKN were also demonstrated.

MKN is a knowledge-based inference model applicable to many AI problems, but leaves ample space for the future. Directions for future work fall into three main areas:

**Knowledge base:** We plan to annotate more records and structure more medical knowledge to investigate the application of MKN in a variety of domains.

**Inference:** We plan to test the effectiveness of the numeric rules after the test data have been satisfied, identifying and exploiting the possibility of inference throughout the knowledge base.

**Learning:** We plan to develop algorithms for learning and replace the pseudo-log-likelihood function, study dynamic approaches to weight learning, and build MKNs from sparse data and incomplete data.

## Acknowledgements

The Chinese Electronic Medical Records used in this paper were provided by the Second Affiliated Hospital of Harbin Medical University. We would like to thank the reviewers for their detailed reviews and insightful comments, which have helped to improve the quality of this paper.

## References

[1] World Health Organization (WHO), "Diabetes Programme". [Online] Available: http://www.who.int/diabetes/en/.

[2] World Health Organization (WHO), "Cardiovascular disease". [Online] Available: http://www.who.int/cardiovascular_diseases/en/.

[3] World Health Organization (WHO), "Cancer". [Online] Available: http://www.who.int/cancer/en/.

[4] A.T. Azar, S.M. El-Metwally, Decision tree classifiers for automated medical diagnosis, Neural Computing and Applications. 23(7-8) (2013) 2387-2403.

[5] D. Lavanya, K.U. Rani, Ensemble decision tree classifier for breast cancer data, International Journal of Information Technology Convergence and Services. 2(1) (2012) 17.

[6] Y.C.T.Bo. Jin, Support vector machines with genetic fuzzy feature transformation for biomedical data classification, Inf Sci. 177(2) (2007) 476-489.

[7] M. Peker, A decision support system to improve medical diagnosis using a combination of k-medoids clustering based attribute weighting and SVM, Journal of medical systems. 40(5) (2016) 1-16.

[8] D. Vassis, B.A. Kampouraki, P. Belsis, et al., Using neural networks and SVMs for automatic medical diagnosis: a comprehensive review, INTERNATIONAL CONFERENCE ON INTEGRATED INFORMATION (IC-ININFO). 1644(1) (2015) 32-36.

[9] Y.Y. Wee, W.P. Cheah, S.C. Tan, et al., A method for root cause analysis with a Bayesian belief network and fuzzy cognitive map, Expert Systems with Applications. 42(1) (2015) 468-487.

[10] A.C. Constantinou, N. Fenton, W. Marsh, et al., From complex questionnaire and interviewing data to intelligent Bayesian network models for medical decision support, Artificial intelligence in medicine. 67(1) (2016) 75-93.

[11] Y. Huang, P. McCullagh, N. Black, R. Harper, Feature selection and classification model construction on type 2 diabetic patients' data, Artif. Intell. Med. 41(3) (2007) 251-262.

[12] X. Liu, R. Lu, J. Ma, et al., Privacy-preserving patient-centric clinical decision support system on naive Bayesian classification, IEEE journal of biomedical and health informatics. 20(2) (2016) 655-668.

[13] A. Bhardwaj, A. Tiwari, Breast cancer diagnosis using genetically optimized neural network model, Expert Systems with Applications. 42(1) (2015) 4611-4620.

[14] F. Amato, A. López, E.M. Peña-Méndez, et al., Artificial neural networks in medical diagnosis, Journal of applied biomedicine. 11(2) (2013) 47-58.

[15] S. Palaniappan, R. Awang, Intelligent heart disease prediction system using data mining techniques, IEEE/ACS International Conference on Computer Systems and Applications. (2008) 108-115.

[16] J.A. Sanz, M. Galar, A. Jurio, et al., Medical diagnosis of cardiovascular diseases using an interval-valued fuzzy rule-based classification system, Applied Soft Computing. 20(1) (2014) 103-111.

[17] N.A. Korenevskiy, Application of Fuzzy Logic for Decision-Making in Medical Expert Systems, Biomedical Engineering. 49(1) (2015) 46-49.

[18] C.A. Pena-Reyes, M. Sipper, Evolutionary computation in medicine: an overview, Artificial Intelligence in Medicine. 19(1) (2000) 1-23.

[19] A. Jain, Medical Diagnosis using Soft Computing Techniques: A Review, International Journal of Artificial Intelligence and Knowledge Discovery. 5(3) (2015) 11-17.

[20] W.P. Goh, X. Tao, J. Zhang, et al., Decision support systems for adoption in dental clinics: a survey, Knowledge-Based Systems. 104(1) (2016) 195-206.

[21] M. Richardson, P. Domingos, Markov logic networks, Machine learning. 62(1-2) (2006) 107-136.

[22] PRIMARY PSYCHIATRY, "Electronic Medical Records." [Online] Available: http://primarypsychiatry.com/electronic-medical-records/.

[23] A.K. Sari, W. Rahayu, M. Bhatt, Archetype sub-ontology: Improving constraint-based clinical knowledge model in electronic health records, Knowledge-Based Systems. 26(1) (2012) 75-85.

[24] S.L. Ting, S.K. Kwok, A.H.C. Tsang, et al., A hybrid knowledge-based approach to supporting the medical prescription for general practitioners: Real case in a Hong Kong

medical center, Knowledge-Based Systems. 24(3) (2011) 444-456.

[25] Ö. Uzuner, Y. Luo, P. Szolovits, Evaluating the state-of-the-art in automatic de-identification, Journal of the American Medical Informatics Association. 14(5) (2007) 550-563.

[26] J. Kazama, K. Torisawa, Exploiting Wikipedia as external knowledge for named entity recognition, Proceedings of the 2007 Joint Conference on Empirical Methods in Natural Language Processing and Computational Natural Language Learning (EMNLP-CoNLL). (2007) 698-707.

[27] M. Song, W.C. Kim, D. Lee, et al., PKDE4J: Entity and relation extraction for public knowledge discovery, Journal of biomedical informatics. 57(1) (2015) 320-332.

[28] I2B2, "Informatics for Integrating Biology & the Bedside." [Online] Avaiable: https://www.i2b2.org/.

[29] WILAB-HIT, "Resources." [Online] Avaiable: https://github.com/WILAB-HIT/Resources/.

[30] J. Pearl, Probabilistic reasoning in intelligent systems: networks of plausible inference, Morgan Kaufmann (2014).

[31] E. Ising, Beitrag zur theorie des ferromagnetismus, Zeitschrift für Physik A Hadrons and Nuclei. 31(1) (1925) 253-258.

[32] G.E. Hinton, T.J. Sejnowski, Optimal perceptual inference, Proceedings of the IEEE conference on Computer Vision and Pattern Recognition. (1983) 448-453.

[33] E. Yilmaz, E. Kanoulas, J.A. Aslam, A simple and efficient sampling method for estimating AP and NDCG, Proceedings of the 31st annual international ACM SIGIR conference on Research and development in information retrieval (2008).

[34] Project Tuffy, "Meet Tuffy." [Online] Avaiable: http://i.stanford.edu/hazy/tuffy/.

**Figure Captions**

**Fig. 1.** Sample of discharge summary from the Second Affiliated Hospital of Harbin Medical University.

**Fig. 2.** Sample of progress note from the Second Affiliated Hospital of Harbin Medical University.

**Fig. 3.** Topology of the knowledge graph.

**Fig. 4.** DCG (discounted cumulative gain) for discharge summaries and progress notes using different measures of quality for the disease node.

**Fig. 5.** Distribution of R@10 (recall for first 10 results) for discharge summaries and progress notes using different measures of quality for the disease node.

**Fig. 6.** DCG (discounted cumulative gain) for discharge summaries and progress notes using different types of symptom variable.

**Fig. 7.** Distribution of R@10 (recall for first 10 results) for discharge summaries and progress notes using different types of symptom variable.

**Fig. 8.** DCG and R@10 for all CEMRs (Chinese Electronic Medical Records): MLN (Markov logic network), MKN (medical knowledge network), and LR (logistic regression algorithm).